\documentclass[letterpaper]{article} 
\usepackage{aaai2027}
\usepackage[hyphens]{url}  
\usepackage{graphicx} 
\usepackage{natbib}  
\usepackage{caption} 
\usepackage{algorithm}
\usepackage{algorithmic}

\usepackage{amssymb}   
\usepackage{bm}        
\usepackage{amsmath}
\usepackage{multirow}
\usepackage[table]{xcolor}  

\usepackage{newfloat}
\usepackage{listings}
\DeclareCaptionStyle{ruled}{labelfont=normalfont,labelsep=colon,strut=off} 
\floatstyle{ruled}
\newfloat{listing}{tb}{lst}{}
\floatname{listing}{Listing}

\usepackage{booktabs}

\title{RSPO: Reward-Swap Policy Optimization for Multi-Turn LLM Agents}

\author{
    Qiang Liu\textsuperscript{\rm 1}\equalcontrib,
    Taian Guo\textsuperscript{\rm 1}\equalcontrib,
    Ruizhi Qiao\textsuperscript{\rm 1}\corresponding,
    Xing Sun\textsuperscript{\rm 1}
}

\affiliations{
    \textsuperscript{\rm 1}Tencent YouTu Lab\\
    Ruizhi.Qiao@tencent.com
}

\begin{document}

\maketitle

\begin{abstract}
Reinforcement learning holds significant potential for training large language models (LLMs) to handle multi-turn interactive tasks. However, in long-horizon, multi-turn tasks characterized by sparse outcome rewards, directly training with outcome rewards often results in slow convergence due to the sparsity of signals and the lack of fine-grained feedback. 
Furthermore, the model may fail to learn successful trajectories that are not sampled during training, thereby limiting its performance. Conversely, while employing customized dense process rewards provides richer signals and accelerates convergence, these surrogate rewards may exhibit potential misalignment with the ground-truth outcome rewards. This inconsistency can bias the training direction and ultimately degrade the model's final performance. In this work, we propose Reward-Swap Policy Optimization (RSPO), a method designed to leverage the rich information from dense process rewards to facilitate training with outcome rewards. By utilizing a reward-swap mechanism, RSPO ensures the diversity of sampled trajectories while guaranteeing consistency between the optimization objective and the true outcome rewards, thereby elevating the performance ceiling of the model. We conduct extensive experiments on two challenging agent benchmarks, WebShop and ALFWorld. By applying our method to various reinforcement learning algorithms, including GRPO, PPO, and GiGPO, we demonstrate that RSPO achieves consistent performance improvements across different baselines and benchmarks.
\end{abstract}


\section{Introduction}

\begin{figure}[t]
    \centering
  \includegraphics[width=\linewidth]{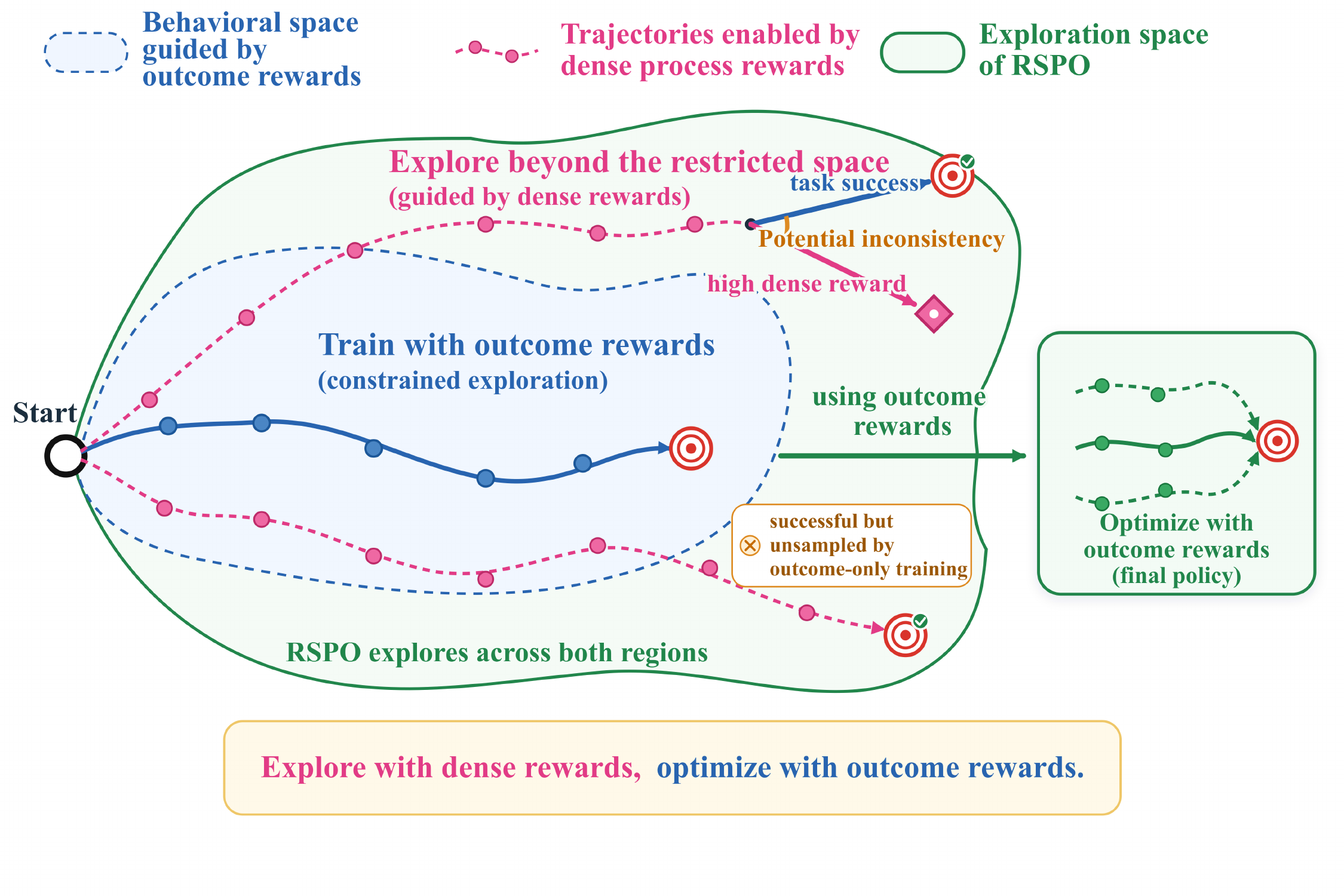}
  \caption{
  Schematic illustration of the motivation for RSPO and its core idea. Training only with sparse outcome rewards can concentrate sampled trajectories within a limited behavioral region and leave some successful trajectories unsampled. Dense process rewards provide step-wise guidance that enables exploration beyond this region, but optimizing them may favor a direction that is inconsistent with task success. RSPO explores across both outcome-reward-guided and dense-rewards-guided regions. Trajectories collected under dense-rewards guidance are stored in the replay buffer and, together with on-policy trajectories, are used to update the final policy with task outcome rewards. Thus, dense rewards guide exploration, whereas outcome rewards determine the final optimization objective.
  }
  \label{fig:diff}
\end{figure}

In recent years, Large Language Models (LLMs) have demonstrated remarkable generalization capabilities across various domains, achieving breakthrough progress particularly in complex tasks such as mathematical problem solving, code generation, and logical reasoning~\cite{wei2022chain, achiam2023gpt, singhal2023large, wu2023bloomberggpt}. To further elicit the deep reasoning capabilities of these models, Reinforcement Learning (RL) has been widely employed to fine-tune LLMs, optimizing the model policy by maximizing the expected reward of the generated content~\cite{ouyang2022training, bai2022constitutional}. Within this mainstream paradigm, researchers typically model reasoning tasks as a single-turn interaction process: the model receives a specific question or instruction as input and generates a Chain-of-Thought (CoT) or final answer in a single pass, after which the environment provides reward feedback based on the quality of the output~\cite{yu2025dapo, liu2025understanding}. 

However, many RL fine-tuning setups for these tasks model each example as a single prompt-response episode.
In contrast, LLM agents frequently encounter multi-turn tasks characterized by outcome-only rewards~\cite{alfworld}. When addressing such tasks, directly employing outcome rewards for reinforcement learning often leads to slow convergence due to the sparsity of reward signals and the absence of fine-grained feedback. Moreover, for algorithms like GRPO~\cite{shao2024deepseekmath}, the model may fail to learn successful trajectories that are not sampled during training, thereby limiting its performance.

To mitigate these challenges, several approaches, such as~\cite{spa, zhang2025rlvmr}, construct or learn dense process rewards to alleviate reward sparsity. By assigning an intermediate reward to each step of the trajectory, these methods provide richer reward signals, resulting in faster model convergence. However, methods relying on customized dense process rewards face a potential problem: if the dense rewards are not perfectly accurate, they may be inconsistent with the task's outcome rewards. This inconsistency can bias the training direction and ultimately degrade the model's final performance.

In this work, we introduce Reward-Swap Policy Optimization (RSPO),
a general reinforcement learning framework that leverages dense process rewards to enhance training with sparse outcome rewards.
As illustrated in Figure~\ref{fig:diff}, training only with sparse outcome rewards can concentrate sampled trajectories within a limited behavioral region and leave some successful trajectories unsampled. Dense process rewards provide step-wise guidance that enables exploration beyond this region, but optimizing them may favor a direction that is inconsistent with task success. RSPO therefore explores across both outcome-reward-guided and dense-rewards-guided regions and incorporates trajectories collected under dense-rewards guidance into final-policy updates performed with task outcome rewards. 
The detailed cyclic implementation—including the branches of Agent $A$ and Agent $B$, reward swapping, sampling from the replay buffer $\mathcal{D}$, the generalized clipping-center shift, and the combination of the on-policy objective $\mathcal{J}_{\text{On}}$ and the off-policy objective $\mathcal{J}_{\text{Off}}$ into $\mathcal{J}_{\text{Total}}$—is illustrated in Figure~\ref{fig:algo_overview}.

Extensive experiments conducted on ALFWorld, an embodied agent environment simulating household tasks, and WebShop, a web search interaction task, demonstrate that applying our RSPO method to various reinforcement learning algorithms—including GRPO~\cite{shao2024deepseekmath}, GiGPO~\cite{gigpo}, and PPO~\cite{schulman2017proximal}—yields significant performance gains. In experiments utilizing Qwen2.5-1.5B-Instruct~\cite{qwen2}, 
RSPO achieved a 5.7\% higher success rate than the GRPO baseline and an 8.6\% higher success rate than the PPO baseline in the ALFWorld environment.
Similarly, in the WebShop environment, RSPO yields a 5.5\% gain over GRPO and a 12\% gain over PPO.
On average across all tasks, RSPO improved success rates by 4.87\% over GRPO, 3.6\% over GiGPO, and 6.6\% over PPO. 
These results demonstrate the effectiveness of RSPO in addressing such multi-turn tasks characterized by outcome rewards.

\section{Related Work}

\subsection{Reinforcement Learning for LLMs}

Reinforcement Learning (RL) is critical for optimizing LLMs\cite{ouyang2022training}, particularly for reasoning tasks\cite{lightman2023let, guo2025deepseek}. However, standard Actor-Critic methods like PPO\cite{schulman2017proximal} incur high computational costs due to the value network. To address this, recent "critic-free" methods estimate advantages using statistical baselines instead. Examples include GRPO\cite{shao2024deepseekmath}, RLOO\cite{ahmadian2024back}, and REINFORCE++\cite{hu2025reinforceefficientrlhfalgorithm}, which utilize group, leave-one-out, or global batch averages, respectively. Other methods, such as ~\cite{li2024remax, yu2025dapo, liu2025understanding, zheng2025group}, follow a similar paradigm. These methods significantly lower training overhead. Furthermore, to improve data usage efficiency, several works have proposed incorporating off-policy data\cite{yan2025learning, li2025repo, liang2025squeeze}, demonstrating that reusing historical trajectories can effectively facilitate model training.

\subsection{Reinforcement Learning for LLM-based Agents}

Recent work applies RL to enhance tool utilization in LLMs\cite{feng2025retool, li2025torl, jin2025search}. Several methods focus on training stability and credit assignment: RAGEN\cite{wang2025ragen} filters low-variance trajectories, while GiGPO\cite{gigpo} and MT-GRPO\cite{wei2025reinforcingmultiturnreasoningllm} utilize turn-level advantages. 
Furthermore, ARPO~\cite{dong2025agentic} employs additional branching at high-entropy tokens, thereby concentrating computational resources on the optimization of high-entropy steps. AEPO~\cite{aepo} further refines the branching strategy and advantage estimation mechanisms of ARPO.
SPEAR~\cite{qin2025learnropestrustwins} accelerates training via self-imitation\cite{oh2018self} and curriculum learning from replay data, while \cite{zhang2025agent} integrate implicit world modeling and self-reflection to guide action selection. IGPO\cite{igpo} introduces an intrinsic reward mechanism by calculating the information gain based on the conditional probability of the ground truth.

In this work, we introduce a reward-swap reinforcement learning mechanism, which leverages diverse trajectories induced by dense process rewards to facilitate training with sparse outcome rewards. Our proposed RSPO is compatible with existing reinforcement learning methods and can work synergistically to achieve superior performance.

\subsection{Utilizing Process Rewards in LLMs}

When an LLM agent engages in multi-turn interactions with an environment and receives an outcome reward only at the end of a trajectory, direct training with outcome rewards often leads to slow convergence due to reward sparsity. To address this, several approaches accelerate learning by constructing dense process rewards. StepAgent~\cite{deng2024novice} generates step-wise rewards by imitating and reflecting upon expert trajectories. PRIME~\cite{cui2025process} employs implicit process reward modeling to continuously optimize the reward model during training, thereby yielding more accurate process rewards. Similarly, ReasonRAG~\cite{liprocess} utilizes Shortest Path Reward Estimation, approximating process rewards via Monte Carlo-style estimation with step-based penalties. SPA-RL~\cite{spa} decomposes the final reward into fine-grained intermediate signals through a step-wise progress attribution mechanism. Building upon the ReAct~\cite{yao2022react} framework, RLVMR~\cite{zhang2025rlvmr} introduces meta-reasoning labels to derive more sophisticated composite rewards. However, while these dense process rewards can accelerate model training, such methods remain susceptible to potential inconsistencies between process rewards and outcome rewards.

\section{Preliminaries}

\subsection{Task setup}

We formulate the multi-turn interaction between the LLM agent and the environment as a Markov Decision Process (MDP)~\cite{RL2018}. For a task $x \sim p(X)$, the agent observes a state $\bm{s}_t \in \mathcal{S}$ at step $t$ and samples an action $\bm{a}_t \sim \pi_\theta(\cdot | \bm{s}_t)$ based on parameters $\theta$. The environment then returns the next state $\bm{s}_{t+1}$ and a reward $\bm{r}_t$. This process yields a trajectory $\bm{\tau} = \{(\bm{s}_t, \bm{a}_t, \bm{r}_t)\}_{t=1}^T$. Reinforcement learning aims to optimize $\theta$ by maximizing the expected discounted return $\mathbb{E}_{\bm{\tau}\sim\pi_\theta}\left[\sum_{t=1}^{T}\gamma^{t-1}\bm{r}_t\right]$, where $\gamma \in (0,1]$ is the discount factor.

\subsection{Group Relative Policy Optimization (GRPO)}

Recently, critic-free methods like GRPO have proven highly effective for LLM optimization. By eliminating the value function required by PPO, GRPO reduces memory and computational resource consumption. For each task $x$, GRPO samples $N$ trajectories $\{\bm{\tau}_1, \dots, \bm{\tau}_N\}$ and normalizes the rewards to compute the advantage:

\begin{equation*}
A(\bm{\tau}_i)=\frac{R(\bm{\tau}_i) - \text{mean}\bigl(\{R(\bm{\tau}_j)\}_{j=1}^N\bigr)}{{\text{std}}\bigl(\{R(\bm{\tau}_j)\}_{j=1}^N\bigr)}.
\end{equation*}

The objective function is formulated as:
\begin{align*}
\label{eq:grpo_objective1}
&\mathcal{J}_{\mathrm{\text{GRPO}}}
=
\mathbb{E}_{\substack{x \sim p(X), \{\bm{\tau}_i\}_{i=1}^N \sim \pi_{\theta_{\text{old}}}}}
\biggl[
    \frac{1}{NT}
    \sum_{i=1}^{N}
    \sum_{t=1}^{T} \notag\\
    &\min\Bigl(
    r_{i,t}(\theta){A}_{i,t},\,
    \text{clip}\bigl(r_{i,t}(\theta), 1 \pm \epsilon\bigr) {A}_{i,t})
    \Bigr)
\biggr] \nonumber \\
&-\beta
    \mathbb{D}_{\mathrm{KL}}\!\bigl(\pi_\theta(\cdot \mid x) \,\|\, \pi_{\mathrm{ref}}(\cdot \mid x)\bigr),
\end{align*}
where $r_{i,t}$ is the importance sampling ratio $\frac{\pi_{\theta}(a_t^{i}\mid x,\,s_t^{i})}{\pi_{\theta_{\mathrm{old}}}(a_t^{i}\mid x,\,s_t^{i})}$. To ensure training stability, the KL divergence term penalizes excessive deviation of the policy $\pi_{\theta}$ from the reference model $\pi_{\mathrm{ref}}$, scaled by the factor $\beta$.

\begin{figure*}[t]
   \begin{center}
   \includegraphics[width=\linewidth]{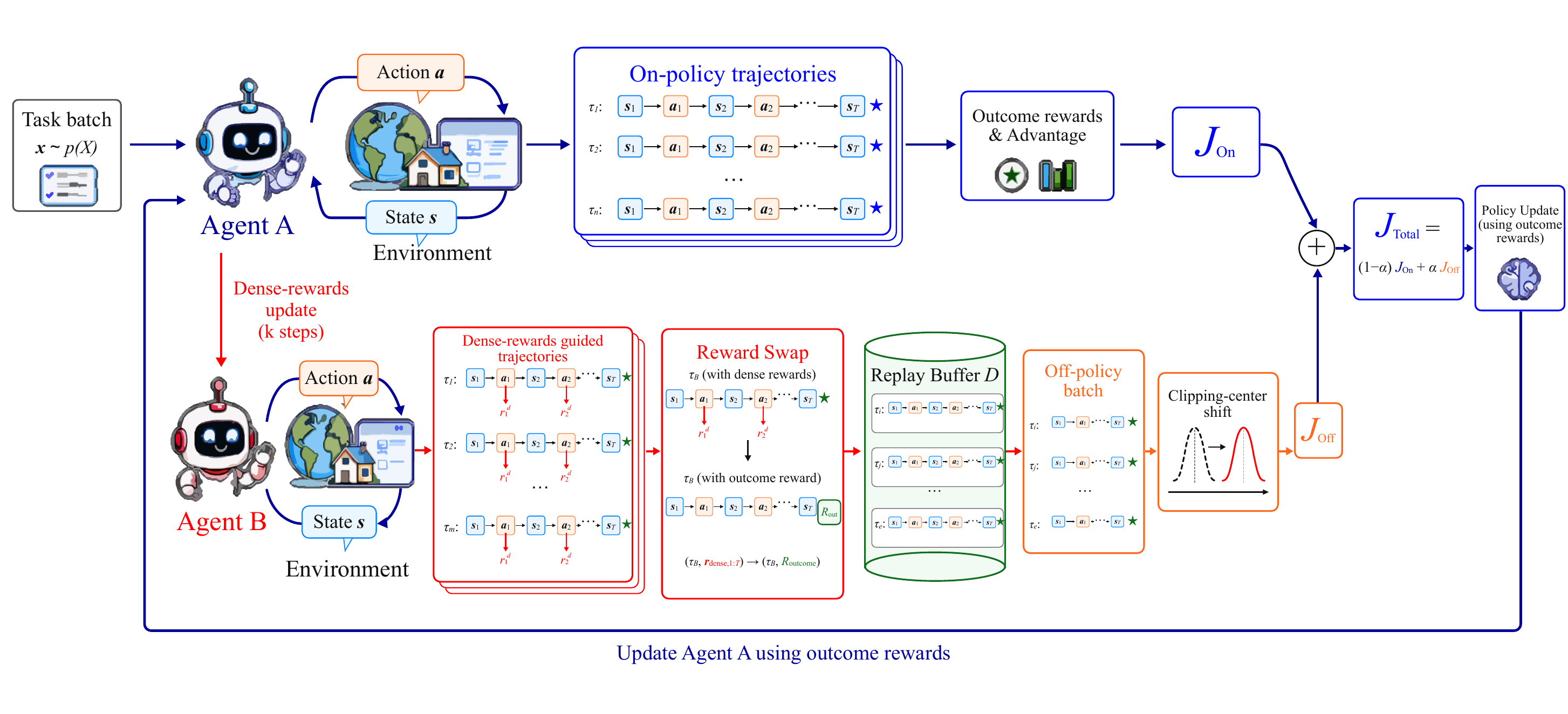}
   \end{center}
   \caption{
   Overview of RSPO. $n$ and $m$ represent the number of on-policy and off-policy data, respectively. First, Agent $A$ is trained using dense process rewards for $k$ steps to obtain Agent $B$. Agent $B$ then interacts with the environment to generate a batch of dense-rewards guided trajectories. Through reward swap, the dense process rewards associated with these trajectories are replaced with the task's outcome rewards before the trajectories are stored in the replay buffer $\mathcal{D}$. For the final-policy update, Agent $A$ generates on-policy trajectories, whose outcome rewards and advantages are used to construct $\mathcal{J}_{\mathrm{On}}$. Meanwhile, an off-policy batch sampled from $\mathcal{D}$ is used to construct $\mathcal{J}_{\mathrm{Off}}$ with the generalized clipping-center shift. The two objectives are combined as $\mathcal{J}_{\mathrm{Total}}=(1-\alpha)\mathcal{J}_{\mathrm{On}}+\alpha \mathcal{J}_{\mathrm{Off}}$ to update Agent $A$ using outcome rewards. The updated Agent $A$ then initializes the next training cycle. Thus, dense rewards guide trajectory exploration, while outcome rewards determine the final policy optimization.
}
   \label{fig:algo_overview}
\end{figure*}

\section{Reward-Swap Policy Optimization}

\subsection{Motivation}

The motivation for our method in addressing multi-turn tasks with sparse outcome rewards stems from the human learning process regarding delayed rewards. When humans are confronted with a task characterized by delayed rewards and only outcome feedback, identifying an effective direction initially can be challenging. To overcome this, we often establish artificial immediate rewards—intermediate signals that we hypothesize will contribute to the final objective, even if they are not perfectly aligned with the task's outcome rewards. We then optimize our choices based on these immediate rewards until certain attempts successfully complete the task and elicit the final outcome feedback. Subsequently, we retrospectively analyze these successful trajectories to reflect and learn, discerning which specific steps genuinely contributed to the long-term goal. By reinforcing these effective steps, we proficiently master the task.

\subsection{RSPO}

Based on this motivation, we derive the overall training pipeline of RSPO. In this framework, the tasks are formulated similarly: they require the agent to engage in multi-turn interactions with the environment, which returns an outcome reward only upon task completion. Here, the immediate rewards correspond to dense process rewards; while beneficial for achieving the final objective, they may exhibit potential inconsistencies with the task's outcome rewards.


Following this rationale, we first train the initial Agent $A$ using dense process rewards for $k$ steps to obtain the intermediate Agent $B$. Agent $B$ then interacts with the environment to collect a batch of dense-rewards-guided trajectories. Through reward swap, their dense process rewards are replaced with task outcome rewards before the trajectories are stored in the replay buffer $\mathcal{D}$. For each final-policy update, Agent $A$ generates on-policy trajectories to construct $\mathcal{J}_{\mathrm{On}}$, while an off-policy batch sampled from $\mathcal{D}$ is used to construct $\mathcal{J}_{\mathrm{Off}}$ with the generalized clipping-center shift. The two objectives are combined to form $\mathcal{J}_{\mathrm{Total}}$, and Agent $A$ is updated using task outcome rewards for another $k$ steps, resulting in an updated Agent $A'$. Finally, Agent $A'$ is used as Agent $A$ in the next iteration, and the cycle repeats. The overall framework is illustrated in Figure~\ref{fig:algo_overview}, and the pseudocode for RSPO is provided in Appendix~D. During the dense-reward phase, PPO uses the step-level scores to compute GAE, whereas GRPO and GiGPO use their summed trajectory return.

When Agent $A$ is updated using the task's outcome rewards, its optimization objective is formulated as follows:
\begin{equation}\label{eq:all}
\mathcal{J}_{\text{Total}}(\pi_\theta)=(1-\alpha)\mathcal{J}_{\text{On}}(\pi_\theta) + \alpha\mathcal{J}_{\text{Off}}(\pi_\theta),
\end{equation}
where $\mathcal{J}_{\text{On}}$ and $\mathcal{J}_{\text{Off}}$ denote the on-policy and off-policy objectives constructed from self-generated trajectories of Agent $A$ and trajectories sampled from the replay buffer $\mathcal{D}$, respectively. Both objectives are computed using task outcome rewards, and $\alpha$ controls the relative contribution of $\mathcal{J}_{\text{Off}}$ to the total objective $\mathcal{J}_{\text{Total}}$. In our implementation, $\alpha$ is set consistently with the proportion of off-policy data in each update batch.

Moreover, since proximal policy gradient methods such as PPO and GRPO typically necessitate on-policy training, an excessive amount of off-policy data may induce a significant deviation in the state-action distribution relative to the current policy, thereby compromising training stability. Consequently, we prioritize the use of self-generated on-policy trajectories during training. In our practical experiments, the value of $\alpha$ is set to $1/8$.

\subsubsection{Sampling Strategy}

When sampling from the replay buffer, the choice of sampling strategy—such as random sampling, variance-based sampling, or reward-based sampling—significantly impacts the quality of the data utilized for updates. Guided by the conclusions from~\cite{li2025repo} and the original insight underlying our motivation (i.e., learn from successful trajectories), we adopt reward-based sampling as our definitive strategy. Specifically, this method prioritizes the sampling of trajectories that yield the highest rewards.

In the meanwhile, to ensure a fair comparison, we guarantee that the amount of data used for training RSPO is consistent with that of the baselines. For example, if the train data size for GRPO training is $16$, RSPO also uses a train data size of $16$, with the data size for self-generated on-policy trajectories and sampled off-policy trajectories being $14$ and $2$, respectively.

\subsubsection{Clipping-center shift}

Proximal policy gradient methods, such as PPO and GRPO, typically employ a clipping operation to prevent excessive policy updates. However, when utilizing off-policy data for updates, a discrepancy exists between the distribution of the current policy and that of the policy used to sample the off-policy data. Directly applying the standard clipping range $(1-\epsilon, 1+\epsilon)$ may result in the erroneous clipping of valid data, rendering their contribution to the gradient update zero. Consequently, for off-policy data, we adopt the Generalized Clipping Mechanism proposed in ~\cite{geppo}:

\begin{equation}\label{eq:geppo}
\text{clip} \left( \frac{\pi_{\theta}(a \mid s)}{\pi_{B}(a \mid s)}, \frac{\pi_{\theta_{\text{old}}}(a \mid s)}{\pi_{B}(a \mid s)} \pm \epsilon\right),
\end{equation}
where $\pi_{\theta}$ and $\pi_{\theta_{\text{old}}}$ represent the current policy being updated and the old policy (i.e., the policy before the update), respectively, and $\pi_B$ represents the policy that generates off-policy data (i.e., the policy corresponding to Agent $B$ in Figure~\ref{fig:algo_overview}).
The Generalized Clipping Mechanism introduces an offset to the clipping center, rendering the constraint more reasonable.
The components of Equation~\ref{eq:all} are provided below:

\begin{align}
\label{eq:gigpo_objective1}
&\mathcal{J}_{\mathrm{\text{On}}}
=
\mathbb{E}_{\substack{x \sim p(X), \{\bm{\tau}_i\}_{i=1}^{N} \sim \pi_{\theta_{\text{old}}}}}
\biggl[
    \frac{1}{NT}
    \sum_{i=1}^{N}
    \sum_{t=1}^{T} \notag\\
    &\min\Bigl(
    r^{\mathrm{on}}_{i,t}(\theta){A}^{\mathrm{on}}_{i,t},\,
    \text{clip}\bigl(r^{\mathrm{on}}_{i,t}(\theta), 1 \pm \epsilon\bigr) {A}^{\mathrm{on}}_{i,t})
    \Bigr)
\biggr] \nonumber \\
&-\beta
    \mathbb{D}_{\mathrm{KL}}\!\bigl(\pi_\theta( \cdot\mid x) \,\|\, \pi_{\mathrm{ref}}( \cdot\mid x)\bigr),
\end{align}
\begin{align}
\label{eq:gigpo_objective}
&\mathcal{J}_{\mathrm{\text{Off}}}
=
\mathbb{E}_{\substack{{x, \tau} \sim \mathcal{D}}}
\biggl[
    \frac{1}{NT}
    \sum_{i=1}^{N}
    \sum_{t=1}^{T} 
    \min\Bigl(
    r^{\mathrm{off}}_{i,t}(\theta)\notag\\
    &{A}^{\mathrm{off}}_{i,t},\,
    \text{clip}\bigl(r^{\mathrm{off}}_{i,t}(\theta), \frac{\pi_{\theta_{\text{old}}}(a \mid s)}{\pi_{B}(a \mid s)} \pm \epsilon\bigr) {A}^{\mathrm{off}}_{i,t})
    \Bigr)
\biggr] \nonumber \\
&-\beta
    \mathbb{D}_{\mathrm{KL}}\!\bigl(\pi_\theta(\cdot \mid x) \,\|\, \pi_{\mathrm{ref}}(\cdot \mid x)\bigr),
\end{align}
where $$r_{i,t}^{\mathrm{on}}
= \frac{\pi_{\theta}\bigl(a_t^{i}\mid x,\,s_t^{i}\bigr)}{\pi_{\theta_{\mathrm{old}}}\bigl(a_t^{i}\mid x,\,s_t^{i}\bigr)}, r_{i,t}^{\mathrm{off}} = \frac{\pi_{\theta}\bigl(a_t^{i}\mid x,\,s_t^{i}\bigr)}{\pi_{B}\bigl(a_t^{i}\mid x,\,s_t^{i}\bigr)}.$$

For GRPO and PPO, advantage estimation employs group-based estimation and the Generalized Advantage Estimator (GAE)~\cite{schulman2016high}, respectively.

\subsubsection{Dense process reward model}

Following SPA-RL~\cite{spa}, we train a dense process reward model using a collected dataset $\mathcal{D}_{dense}=\bigl\{(\hat{\tau}_{i},\,R_{i})\bigr\}_{i=1}^{M}$, where $R_{i}$ is the final reward for trajectory $\hat{\tau}_{i}$. We incorporate a Multi-Layer Perceptron (MLP) after the final hidden layer of the pre-trained LLM $\pi_{pre}$, utilizing a tanh activation to constrain the intermediate reward $r$ within $(-1, 1)$. The formulation is given by:
\begin{equation}
\hat{r}_{t}
= \tanh(\mathrm{MLP}\bigl(h_t\bigr)), 
\quad
h_t = f_{\pi_{pre}}(s_t, a_t),
\end{equation}
where $s_t$ is the context at step $t$, 
$f_{\pi_{pre}}$ denotes the encoding operation of $\pi_{pre}$,
and $\hat{r}_t$ is the step-level reward. The total predicted reward for a trajectory is aggregated as:
\begin{equation}
    \hat{R}
    = \sum_{t=1}^{T} \hat{r}_{t},
\end{equation}
where $T$ denotes the total number of steps in trajectory $\tau$.
The model is optimized by minimizing the MSE loss:
\begin{equation}
\mathcal{L}_{\text{P}}
= \frac{1}{M}
  \sum_{i=1}^{M}
  \bigl(\hat{R}_{i} - R_{i}\bigr)^2,
\end{equation}
where $\hat{R_i}$ is the predicted trajectory return and $R_i \in [0, 1]$ is the task outcome reward. 
It is important to note that the dense process rewards need not be derived exclusively through the aforementioned method; we merely adopted one of many available approaches to obtain them.

\section{Experiments}

In this section, we evaluate the performance of RSPO when applied to various reinforcement learning algorithms. We aim to analyze the effectiveness of RSPO from the following perspectives: (1) the performance improvements achieved by RSPO across various benchmarks when integrated with different reinforcement learning methods; (2) ablation study on sampling strategies; (3) the contribution of trajectories induced by dense rewards to performance improvement; (4) the potential inconsistency of dense rewards.

\subsection{Experimental Setup}\label{detail}

\paragraph{Environments.} We evaluated our method on two agent benchmarks: 

(1) ALFWorld~\cite{alfworld}: An embodied control benchmark simulating a household environment. By leveraging an interactive TextWorld interface, it aligns textual abstract environments with embodied physical environments, enabling agents to complete specified tasks via text instructions. It primarily encompasses six household task categories: Pick \& Place (Pick), Examine in Light (Look), Clean \& Place (Clean), Heat \& Place (Heat), Cool \& Place (Cool), and Pick Two \& Place (Pick2).

(2) WebShop~\cite{webshop}: A web search and interaction benchmark. In this environment, the agent interacts with a simulated shopping website to search for and select products, ultimately executing a "buy now" action. The environment subsequently provides a final reward based on the extent to which the selected product matches the specified requirements.

\paragraph{Agent models.} We use Qwen2.5-1.5B-Instruct and Qwen2.5-7B-Instruct~\cite{qwen2} as our base models. 
For the dense reward model, we use Llama-3.2-3B-Instruct as the base model.

\paragraph{Baselines.} We evaluate RSPO across three common and representative reinforcement learning algorithms: PPO, GRPO, and GiGPO. 
For GRPO and GiGPO, the dense reward is calculated as the sum of the predicted values over the entire trajectory.
Specifically, GiGPO is categorized into two variants based on whether the standard deviation (std) is excluded from step-level advantage
calculation: GiGPO\textsubscript{w/ std} and GiGPO\textsubscript{w/o std}. 
We adopt the best-performing GiGPO variants for each environment as reported in \cite{gigpo}: GiGPO\textsubscript{w/ std} for ALFWorld and GiGPO\textsubscript{w/o std} for WebShop.

Furthermore, we compared our approach against SPEAR combined with GRPO and GiGPO (SPEAR+GRPO and SPEAR+GiGPO; note that SPEAR was not applied to PPO). It is worth noting that the original SPEAR experiments utilized larger training data size and total epochs. To ensure a fair comparison, we reproduced the SPEAR results by aligning these hyperparameters with the settings used in our experiments.

\begin{table*}[ht]
    \centering
\setlength{\tabcolsep}{1.4mm}
        \begin{tabular}{lccccccc|cc}
        \toprule
        \multirow{2}{*}{Method} & \multicolumn{7}{c|}{\textbf{ALFWorld}} & \multicolumn{2}{c}{\textbf{WebShop}} \\
        & Pick & Look & Clean & Heat & Cool & Pick2 & All (\textcolor{teal}{$\Delta$}) & Score & SR (\textcolor{teal}{$\Delta$})\\
        \midrule
        \multicolumn{10}{l}{\textit{Qwen2.5-1.5B-Instruct}} \\
        GRPO & 81.3\textsubscript{\textpm7.4} & 58.1\textsubscript{\textpm23.7} & 85.8\textsubscript{\textpm5.9} & 59.5\textsubscript{\textpm9.2} & 68.1\textsubscript{\textpm13.5} & 41.0\textsubscript{\textpm15.5} & 69.0\textsubscript{\textpm7.1}& 83.3\textsubscript{\textpm1.4} & 65.1\textsubscript{\textpm5.9}\\
        \rowcolor{gray!15}SPEAR+GRPO & 62.5\textsubscript{\textpm11.3} & 50.5\textsubscript{\textpm15.6} & 64.6\textsubscript{\textpm3.9} & 35.9\textsubscript{\textpm14.7} & 37.9\textsubscript{\textpm14.6} & 34.6\textsubscript{\textpm5.3} & 50.0\textsubscript{\textpm2.1}& 65.4\textsubscript{\textpm7.7}& 46.1\textsubscript{\textpm5.1} \\
        \rowcolor{blue!15}RSPO+GRPO & 81.0\textsubscript{\textpm5.8} & 64.6\textsubscript{\textpm9.3} & 81.5\textsubscript{\textpm8.6} & 73.1\textsubscript{\textpm15.8} & 74.7\textsubscript{\textpm3.2} & 65.9\textsubscript{\textpm4.3} & 74.7\textsubscript{\textpm1.2}{\tiny \textbf{\textcolor{teal}{(+5.7\%)}}}& 85.8\textsubscript{\textpm0.5}& 70.6\textsubscript{\textpm4.3}{\tiny \textbf{\textcolor{teal}{(+5.5\%)}}}\\
        GiGPO & 98.2\textsubscript{\textpm3.0} & 70.2\textsubscript{\textpm11.4} & 96.3\textsubscript{\textpm0.7} & 86.5\textsubscript{\textpm5.6} & 84.1\textsubscript{\textpm8.8} & 79.2\textsubscript{\textpm13.0} & 88.8\textsubscript{\textpm1.6}& 82.5\textsubscript{\textpm6.7}& 67.4\textsubscript{\textpm7.8}\\
        \rowcolor{gray!15}SPEAR+GiGPO & 86.1\textsubscript{\textpm7.9} & 61.6\textsubscript{\textpm14.3} & 83.7\textsubscript{\textpm6.8} & 57.8\textsubscript{\textpm9.5} & 66.8\textsubscript{\textpm1.6} & 53.5\textsubscript{\textpm17.4} & 71.9\textsubscript{\textpm4.3}& 69.2\textsubscript{\textpm11.6}& 44.5\textsubscript{\textpm9.0}\\
        \rowcolor{blue!15}RSPO+GiGPO & 92.9\textsubscript{\textpm3.7} & 85.4\textsubscript{\textpm4.9} & 97.4\textsubscript{\textpm2.3} & 94.9\textsubscript{\textpm5.0} & 89.8\textsubscript{\textpm5.9} & 73.6\textsubscript{\textpm17.9} & 90.4\textsubscript{\textpm1.2}{\tiny \textbf{\textcolor{teal}{(+1.6\%)}}}& 88.2\textsubscript{\textpm2.4}& 75.3\textsubscript{\textpm3.9}{\tiny \textbf{\textcolor{teal}{(+7.9\%)}}}\\
        PPO & 78.5\textsubscript{\textpm9.3} & 67.7\textsubscript{\textpm4.6} & 77.7\textsubscript{\textpm7.4} & 66.1\textsubscript{\textpm7.9} & 70.7\textsubscript{\textpm9.1} & 38.0\textsubscript{\textpm3.1} & 68.0\textsubscript{\textpm7.5}& 74.4\textsubscript{\textpm8.3}& 57.3\textsubscript{\textpm4.6}\\
        \rowcolor{blue!15}RSPO+PPO & 83.0\textsubscript{\textpm2.8} & 56.1\textsubscript{\textpm6.9} & 88.8\textsubscript{\textpm2.1} & 74.6\textsubscript{\textpm4.5} & 80.0\textsubscript{\textpm8.8} & 58.9\textsubscript{\textpm7.8} & 76.6\textsubscript{\textpm2.8}{\tiny \textbf{\textcolor{teal}{(+8.6\%)}}}& 84.3\textsubscript{\textpm2.1}& 69.3\textsubscript{\textpm6.6}{\tiny \textbf{\textcolor{teal}{(+12.0\%)}}}\\
        \midrule
        \multicolumn{10}{l}{\textit{Qwen2.5-7B-Instruct}} \\
        GRPO & 92.3\textsubscript{\textpm8.3} & 73.2\textsubscript{\textpm9.9} & 92.2\textsubscript{\textpm5.2} & 67.8\textsubscript{\textpm2.6} & 71.7\textsubscript{\textpm6.4} & 54.0\textsubscript{\textpm14.5} & 78.4\textsubscript{\textpm3.7}& 78.3\textsubscript{\textpm6.0}& 68.5\textsubscript{\textpm6.3}\\
        \rowcolor{gray!15}SPEAR+GRPO & 91.9\textsubscript{\textpm6.6} & 58.6\textsubscript{\textpm7.0} & 79.6\textsubscript{\textpm9.4} & 59.7\textsubscript{\textpm21.3} & 54.4\textsubscript{\textpm8.8} & 40.0\textsubscript{\textpm4.8} & 68.5\textsubscript{\textpm7.1}& 76.7\textsubscript{\textpm6.7}& 61.5\textsubscript{\textpm9.3} \\
        \rowcolor{blue!15}RSPO+GRPO & 91.4\textsubscript{\textpm5.6} & 68.2\textsubscript{\textpm20.8} & 94.4\textsubscript{\textpm7.1} & 74.6\textsubscript{\textpm8.4} & 70.0\textsubscript{\textpm3.5} & 60.1\textsubscript{\textpm10.7} & 80.2\textsubscript{\textpm1.8}{\tiny \textbf{\textcolor{teal}{(+1.8\%)}}}& 84.8\textsubscript{\textpm2.5}& 75.0\textsubscript{\textpm1.6}{\tiny \textbf{\textcolor{teal}{(+6.5\%)}}}\\
        GiGPO & 94.7\textsubscript{\textpm2.3} & 82.3\textsubscript{\textpm9.1} & 100.0\textsubscript{\textpm0.0} & 91.6\textsubscript{\textpm2.7} & 83.7\textsubscript{\textpm5.6} & 92.4\textsubscript{\textpm2.1} & 92.4\textsubscript{\textpm1.2}& 86.8\textsubscript{\textpm2.7}& 75.3\textsubscript{\textpm3.0}\\
        \rowcolor{gray!15}SPEAR+GiGPO & 88.4\textsubscript{\textpm6.1} & 76.3\textsubscript{\textpm11.0} & 88.7\textsubscript{\textpm4.2} & 69.6\textsubscript{\textpm13.0} & 62.0\textsubscript{\textpm6.0} & 78.1\textsubscript{\textpm14.4} & 78.4\textsubscript{\textpm4.0}& 83.5\textsubscript{\textpm2.3}& 67.7\textsubscript{\textpm3.6}\\
        \rowcolor{blue!15}RSPO+GiGPO & 98.9\textsubscript{\textpm2.0} & 88.4\textsubscript{\textpm4.4} & 100.0\textsubscript{\textpm0.0} & 100.0\textsubscript{\textpm0.0} & 85.1\textsubscript{\textpm3.8} & 91.3\textsubscript{\textpm7.8} & 95.3\textsubscript{\textpm0.8}{\tiny \textbf{\textcolor{teal}{(+2.9\%)}}}& 89.0\textsubscript{\textpm2.2}& 77.3\textsubscript{\textpm3.1}{\tiny \textbf{\textcolor{teal}{(+2.0\%)}}}\\
        PPO & 89.3\textsubscript{\textpm8.6} & 82.3\textsubscript{\textpm9.1} & 90.0\textsubscript{\textpm9.0} & 83.1\textsubscript{\textpm2.7} & 70.3\textsubscript{\textpm7.5} & 61.8\textsubscript{\textpm9.9} & 80.7\textsubscript{\textpm3.7}& 84.7\textsubscript{\textpm2.4}& 70.6\textsubscript{\textpm3.3}\\
        \rowcolor{blue!15}RSPO+PPO & 94.5\textsubscript{\textpm6.8} & 70.2\textsubscript{\textpm14.6} & 94.7\textsubscript{\textpm4.7} & 76.3\textsubscript{\textpm5.5} & 77.2\textsubscript{\textpm7.3} & 66.9\textsubscript{\textpm8.1} & 83.1\textsubscript{\textpm3.2}{\tiny \textbf{\textcolor{teal}{(+2.4\%)}}}& 85.3\textsubscript{\textpm2.8}& 74.0\textsubscript{\textpm4.4}{\tiny \textbf{\textcolor{teal}{(+3.4\%)}}}\\
        \bottomrule
        \end{tabular}
        \caption{Performance on ALFWorld and WebShop (\%). Values are mean ± standard deviation over three seeds (0, 200, and 400). SR: Success Rate.
    }
    \label{tab:main}
\end{table*}

\paragraph{Training Details.} Following the GiGPO codebase~\cite{gigpo}, we set the train data size to $16$ and rollout group size $N=8$. For RSPO-based methods, the train data size comprises $14$ on-policy and $2$ off-policy data size. Consequently, the total batch size for PPO is $128 (16 \times 8)$, and for RSPO-PPO is $112 (14 \times 8)$. In each RSPO loop, both Agent $A$ and Agent $B$ are updated for $3$ steps respectively. All experiments run for a total of $150$ steps (counted based on the update steps of the final obtained model). Full training settings and hyperparameter details are provided in Appendix~A. In the meanwhile, Appendix~B and C present additional experiments, including hyperparameter sweeps and robustness tests against dense reward noise.

\subsection{Performance}

Table~\ref{tab:main} demonstrates the superior performance of RSPO.
As observed, RSPO demonstrates consistent performance gains across both 1.5B and 7B model scales, with improvements reaching up to 12\%. 
Specifically, using the 1.5B base model, RSPO has a 5.7\% higher success rate than GRPO on ALFWorld and 5.5\% higher on WebShop. 
Compared to GiGPO, RSPO improves success rates by 1.6\% on ALFWorld and 7.9\% on WebShop. The improvement is even more noticeable compared to PPO, with success rates up by 8.6\% on ALFWorld and 12\% on WebShop.

With the 7B base model, the overall performance gains are less substantial compared to the 1.5B model; however, it still achieves average improvements of 2.37\% and 3.96\% on ALFWorld and WebShop, respectively. We attribute this to the stronger foundational capabilities of the 7B model compared to its 1.5B counterpart, which enable it to spontaneously plan and select a greater diversity of trajectories. Consequently, its reliance on the diverse trajectories induced by the dense reward model is diminished, as the model itself is capable of generating a sufficient variety of successful trajectories. This intrinsic capability results in performance improvements that are less pronounced than those observed with the 1.5B model.

Additionally, we reproduce the performance of SPEAR+GRPO and SPEAR+GiGPO under the same settings for training data size and total epochs; however, the results were suboptimal. 
We attribute this to the fact that the original SPEAR paper uses a training data size of 32 and 350 total epochs, while our experiments use a data size of 16 and 150 epochs. Given SPEAR's inherently slower learning speed, it yields suboptimal performance.

\subsection{Ablation Study}
\begin{table}[htp]
  \centering
  \begin{tabular}{lcc}
    \toprule
    Method & Score & SR \\
    \midrule
    Random & 25.8\textsubscript{\textpm13.2} & 12.8\textsubscript{\textpm6.1} \\
    Variance & 78.5\textsubscript{\textpm3.2} & 63.0\textsubscript{\textpm3.3} \\
    Reward & \textbf{85.8}\textsubscript{\textpm0.5}& \textbf{70.6}\textsubscript{\textpm4.3} \\
    \midrule
    $\text{RSPO}_{H}$ & 73.9\textsubscript{\textpm0.5} & 57.0\textsubscript{\textpm4.7} \\
    $\text{RSPO}_{O}$ & 82.5\textsubscript{\textpm6.6} & 64.1\textsubscript{\textpm4.7} \\
    RSPO & \textbf{85.8}\textsubscript{\textpm0.5}& \textbf{70.6}\textsubscript{\textpm4.3} \\
    \bottomrule
  \end{tabular}
      \caption{Ablation results on WebShop (\%). Values are mean ± standard deviation over three seeds (0, 200, and 400).}
  \label{tab:abl}
\end{table}

In WebShop, we conduct ablation studies on the replay buffer's sampling strategies and data sources to validate the efficacy of our reward-based sampling mechanism and the trajectories induced by dense rewards. For all experiments, we employ RSPO+GRPO as the baseline method with Qwen2.5-1.5B-Instruct as the base model. The comprehensive results are presented in Table~\ref{tab:abl}.

\subsubsection{Sampling Strategy}

We evaluated three distinct strategies for sampling from the replay buffer:

(1) Random Sampling: Samples are selected uniformly at random from the replay buffer.

(2) Variance-based Sampling: This method prioritizes samples exhibiting the highest intra-group reward variance, under the hypothesis that samples with higher variance provide richer information.

(3) Reward-based Sampling: This method prioritizes samples with the highest rewards, aiming to focus learning on the diverse successful trajectories induced by dense rewards. This is the final method adopted in our work.

As shown in the upper section of Table~\ref{tab:abl}, reward-based sampling significantly outperforms alternative sampling methods. This indicates that the agent effectively acquires superior performance from the diverse successful trajectories induced by dense rewards, which aligns with our motivation to learn by reviewing successful trajectories.

In addition, we observed that the model performs poorly with random sampling, 
exhibiting severe fluctuations in the success rate curve during training.
We attribute this to the fact that, at the early stage of training, the data collected by the agent is mainly negative samples. 
Therefore, random sampling results in off-policy data that is also largely negative. These negative samples compromise training stability—particularly in off-policy scenarios—thereby degrading final performance~\cite{gao2025soft}.

\subsubsection{Data Sources for the Replay Buffer}

Prior works such as ~\cite{li2025repo, liang2025squeeze, qin2025learnropestrustwins}
employed the model's own historical trajectories for off-policy training. We evaluate a similar setup by populating the replay buffer with Agent $A$'s self-generated historical trajectories ($\text{RSPO}_{H}$). 
Additionally, we trained the intermediate policy with outcome rewards instead of dense process rewards ($\text{RSPO}_{O}$, while maintaining other procedures unchanged) to validate the performance gains attributed to dense rewards.

As shown in the bottom section of Table~\ref{tab:abl}, the final performance when training with self-generated historical trajectories is significantly inferior to that of our method. Similarly, replacing dense rewards with outcome rewards also resulted in a performance decline.
This suggests that dense-rewards-induced trajectories are beneficial because they introduce behaviors insufficiently covered by outcome-only rollouts, thereby broadening effective exploration through replay.
Furthermore, we observed that when the amount of training data is held constant, using the model's own historical trajectories does not outperform the original baseline. Its primary advantage lies in improving data usage efficiency rather than final performance.

\subsection{Inconsistency of Dense Rewards}

To demonstrate that the improvements achieved by RSPO are not merely a consequence of the high quality of the dense reward model itself, we also evaluate the performance of training GRPO exclusively with dense rewards in WebShop (using Qwen2.5-1.5B-Instruct as the base model). 

\begin{figure}
\centering
  \includegraphics[width=0.9\columnwidth]{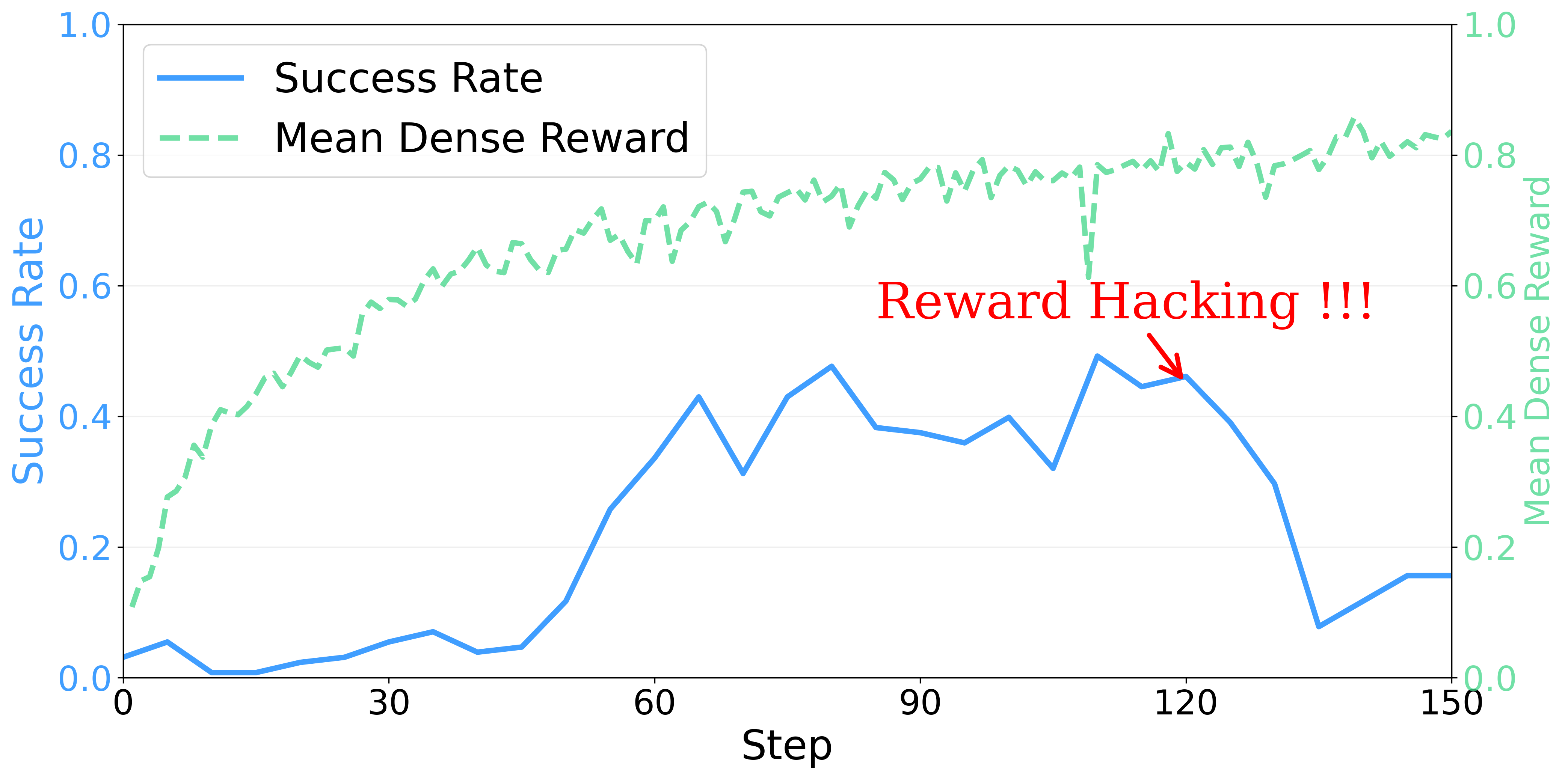}
  \caption{Experimental results of training exclusively with dense rewards. The blue curve represents the task success rate, while the green curve indicates the average dense rewards during training.}
  \label{fig:onlydense}
\end{figure}

The results are illustrated in the Figure~\ref{fig:onlydense}. As observed, when training relies solely on dense rewards, the model's success rate (blue curve) initially increases but subsequently declines, whereas the average dense reward of the sampled trajectories during training (green curve) exhibits a steady upward trend. This indicates that the model focuses on optimizing the dense reward metric; however, since the dense rewards are not perfectly aligned with the task's outcome rewards, this leads to the phenomenon of "Reward Hacking"~\cite{amodei2016concrete}. This finding underscores the necessity of the RSPO, which consistently utilizes outcome rewards for the final model training.

\subsection{Impact of Dense Rewards on Exploration}

To investigate how dense rewards contribute to performance, we analyzed the diversity of visited states during training. We counted the number of identical states visited by the agent
in each training session, the result is shown in Table~\ref{tab:state}.
A higher count indicates the agent is stuck in local optima or repeating behaviors, while a lower count implies better exploration of the state space.

\begin{table}[htp]
  \centering
    \begin{tabular}{lccc}
        \toprule
        \multirow{2}{*}{Method} & \multicolumn{3}{c}{Training Steps} \\
        \cmidrule(lr){2-4}
         & 50 & 100 & 145 \\
        \midrule
        GRPO & 1011 & 1084 & 1167 \\
        \textbf{RSPO} & \textbf{917} & \textbf{889} & \textbf{623} \\
        \bottomrule
    \end{tabular}
      \caption{Comparison of identical state counts within a fixed window during training. Lower values indicate better exploration diversity.}
  \label{tab:state}
\end{table}

As the training progresses, GRPO tends to revisit the same states (count increases to 1167), suggesting it is stuck in a local behavioral space. In contrast, RSPO significantly reduces the number of identical states (dropping to 623), demonstrating that Agent B successfully guides the model to explore a much broader and more diverse state space.

\section{Conclusions}

In this paper, we propose RSPO, a reinforcement learning framework that leverages the rich information from dense process rewards to facilitate training with sparse outcome rewards. This approach addresses both the potential inconsistency of dense process rewards and the sparsity of outcome rewards. Through an iterative process, RSPO enables a model trained with dense process rewards to interact with the environment, generating a wider variety of trajectories. These trajectories are stored in a replay buffer and utilized to assist in training with sparse outcome rewards.

By ensuring that the final model is consistently trained using the task's outcome rewards, RSPO avoids the issues arising from the potential inconsistency of dense process rewards.
Simultaneously, utilizing data from the replay buffer expands the model's exploration space, ultimately elevating its performance ceiling. Experimental results in complex agent environments demonstrate that RSPO yields consistent performance improvements across various reinforcement learning methods, providing a plug-and-play training paradigm for multi-turn LLM agents.

\newpage

\bibliography{aaai2027}

@article{spa,
  title={Spa-rl: Reinforcing llm agents via stepwise progress attribution},
  author={Wang, Hanlin and Leong, Chak Tou and Wang, Jiashuo and Wang, Jian and Li, Wenjie},
  journal={arXiv preprint arXiv:2505.20732},
  year={2025}
}

@book{RL2018,
  title={Reinforcement learning: An introduction},
  author={Sutton, Richard S and Barto, Andrew G and others},
  volume={1},
  number={1},
  year={1998},
  publisher={MIT press Cambridge}
}

@article{geppo,
  title={Generalized proximal policy optimization with sample reuse},
  author={Queeney, James and Paschalidis, Yannis and Cassandras, Christos G},
  journal={Advances in Neural Information Processing Systems},
  volume={34},
  pages={11909--11919},
  year={2021}
}

@article{alfworld,
  title={Alfworld: Aligning text and embodied environments for interactive learning},
  author={Shridhar, Mohit and Yuan, Xingdi and C{\^o}t{\'e}, Marc-Alexandre and Bisk, Yonatan and Trischler, Adam and Hausknecht, Matthew},
  journal={arXiv preprint arXiv:2010.03768},
  year={2020}
}

@article{webshop,
  title={Webshop: Towards scalable real-world web interaction with grounded language agents},
  author={Yao, Shunyu and Chen, Howard and Yang, John and Narasimhan, Karthik},
  journal={Advances in Neural Information Processing Systems},
  volume={35},
  pages={20744--20757},
  year={2022}
}

@article{qwen2,
    title   = {Qwen2.5 Technical Report}, 
    author  = {An Yang and Baosong Yang and Beichen Zhang and Binyuan Hui and Bo Zheng and Bowen Yu and Chengyuan Li and Dayiheng Liu and Fei Huang and Haoran Wei and Huan Lin and Jian Yang and Jianhong Tu and Jianwei Zhang and Jianxin Yang and Jiaxi Yang and Jingren Zhou and Junyang Lin and Kai Dang and Keming Lu and Keqin Bao and Kexin Yang and Le Yu and Mei Li and Mingfeng Xue and Pei Zhang and Qin Zhu and Rui Men and Runji Lin and Tianhao Li and Tingyu Xia and Xingzhang Ren and Xuancheng Ren and Yang Fan and Yang Su and Yichang Zhang and Yu Wan and Yuqiong Liu and Zeyu Cui and Zhenru Zhang and Zihan Qiu},
    journal = {arXiv preprint arXiv:2412.15115},
    year    = {2024}
}

@article{gigpo,
  title={Group-in-group policy optimization for llm agent training},
  author={Feng, Lang and Xue, Zhenghai and Liu, Tingcong and An, Bo},
  journal={Advances in Neural Information Processing Systems},
  volume={38},
  pages={46375--46408},
  year={2026}
}

@inproceedings{lightman2023let,
  title={Let's verify step by step},
  author={Lightman, Hunter and Kosaraju, Vineet and Burda, Yuri and Edwards, Harrison and Baker, Bowen and Lee, Teddy and Leike, Jan and Schulman, John and Sutskever, Ilya and Cobbe, Karl},
  booktitle={International Conference on Learning Representations},
  volume={2024},
  pages={39578--39601},
  year={2024}
}

@article{shao2024deepseekmath,
  title={Deepseekmath: Pushing the limits of mathematical reasoning in open language models},
  author={Shao, Zhihong and Wang, Peiyi and Zhu, Qihao and Xu, Runxin and Song, Junxiao and Bi, Xiao and Zhang, Haowei and Zhang, Mingchuan and Li, YK and Wu, Yang and others},
  journal={arXiv preprint arXiv:2402.03300},
  year={2024}
}

@article{ouyang2022training,
  title={Training language models to follow instructions with human feedback},
  author={Ouyang, Long and Wu, Jeffrey and Jiang, Xu and Almeida, Diogo and Wainwright, Carroll and Mishkin, Pamela and Zhang, Chong and Agarwal, Sandhini and Slama, Katarina and Ray, Alex and others},
  journal={Advances in Neural Information Processing Systems},
  volume={35},
  pages={27730--27744},
  year={2022}
}

@article{schulman2017proximal,
  title={Proximal policy optimization algorithms},
  author={Schulman, John and Wolski, Filip and Dhariwal, Prafulla and Radford, Alec and Klimov, Oleg},
  journal={arXiv preprint arXiv:1707.06347},
  year={2017}
}

@article{guo2025deepseek,
  title={Deepseek-r1: Incentivizing reasoning capability in llms via reinforcement learning},
  author={Guo, Daya and Yang, Dejian and Zhang, Haowei and Song, Junxiao and Zhang, Ruoyu and Xu, Runxin and Zhu, Qihao and Ma, Shirong and Wang, Peiyi and Bi, Xiao and others},
  journal={arXiv preprint arXiv:2501.12948},
  year={2025}
}

@inproceedings{ahmadian2024back,
  title={Back to basics: Revisiting REINFORCE-style optimization for learning from human feedback in LLMs},
  author={Ahmadian, Arash and Cremer, Chris and Gall{\'e}, Matthias and Fadaee, Marzieh and Kreutzer, Julia and Pietquin, Olivier and {\"U}st{\"u}n, Ahmet and Hooker, Sara},
  booktitle={Proceedings of the 62nd Annual Meeting of the Association for Computational Linguistics (Volume 1: Long Papers)},
  pages={12248--12267},
  year={2024}
}

@article{hu2025reinforceefficientrlhfalgorithm,
  title={Reinforce++: An efficient rlhf algorithm with robustness to both prompt and reward models},
  author={Hu, Jian and Liu, Jason Klein and Xu, Haotian and Shen, Wei},
  journal={arXiv preprint arXiv:2501.03262},
  volume={1},
  number={3},
  pages={5},
  year={2025}
}

@article{li2024remax,
  title={Remax: A simple, effective, and efficient reinforcement learning method for aligning large language models},
  author={Li, Ziniu and Xu, Tian and Zhang, Yushun and Lin, Zhihang and Yu, Yang and Sun, Ruoyu and Luo, Zhi-Quan},
  journal={arXiv preprint arXiv:2310.10505},
  year={2023}
}

@article{yu2025dapo,
  title={Dapo: An open-source llm reinforcement learning system at scale},
  author={Yu, Qiying and Zhang, Zheng and Zhu, Ruofei and Yuan, Yufeng and Zuo, Xiaochen and Yue, Yu and Dai, Weinan and Fan, Tiantian and Liu, Gaohong and Liu, Lingjun and others},
  journal={arXiv preprint arXiv:2503.14476},
  year={2025}
}

@article{liu2025understanding,
  title={Understanding r1-zero-like training: A critical perspective},
  author={Liu, Zichen and Chen, Changyu and Li, Wenjun and Qi, Penghui and Pang, Tianyu and Du, Chao and Lee, Wee Sun and Lin, Min},
  journal={arXiv preprint arXiv:2503.20783},
  year={2025}
}

@article{zheng2025group,
  title={Group sequence policy optimization},
  author={Zheng, Chujie and Liu, Shixuan and Li, Mingze and Chen, Xiong-Hui and Yu, Bowen and Gao, Chang and Dang, Kai and Liu, Yuqiong and Men, Rui and Yang, An and others},
  journal={arXiv preprint arXiv:2507.18071},
  year={2025}
}

@article{yan2025learning,
  title={Learning to reason under off-policy guidance},
  author={Yan, Jianhao and Li, Yafu and Hu, Zican and Wang, Zhi and Cui, Ganqu and Qu, Xiaoye and Cheng, Yu and Zhang, Yue},
  journal={arXiv preprint arXiv:2504.14945},
  year={2025}
}

@article{li2025repo,
  title={RePO: Replay-Enhanced Policy Optimization},
  author={Li, Siheng and Zhou, Zhanhui and Lam, Wai and Yang, Chao and Lu, Chaochao},
  journal={arXiv preprint arXiv:2506.09340},
  year={2025}
}

@article{liang2025squeeze,
  title={Squeeze the soaked sponge: Efficient off-policy reinforcement finetuning for large language model},
  author={Liang, Jing and Tang, Hongyao and Ma, Yi and Liu, Jinyi and Zheng, Yan and Hu, Shuyue and Bai, Lei and Hao, Jianye},
  journal={arXiv preprint arXiv:2507.06892},
  year={2025}
}

@article{feng2025retool,
  title={Retool: Reinforcement learning for strategic tool use in llms},
  author={Feng, Jiazhan and Huang, Shijue and Qu, Xingwei and Zhang, Ge and Qin, Yujia and Zhong, Baoquan and Jiang, Chengquan and Chi, Jinxin and Zhong, Wanjun},
  journal={arXiv preprint arXiv:2504.11536},
  year={2025}
}

@article{li2025torl,
  title={Torl: Scaling tool-integrated rl},
  author={Li, Xuefeng and Zou, Haoyang and Liu, Pengfei},
  journal={arXiv preprint arXiv:2503.23383},
  year={2025}
}

@article{jin2025search,
  title={Search-r1: Training llms to reason and leverage search engines with reinforcement learning},
  author={Jin, Bowen and Zeng, Hansi and Yue, Zhenrui and Yoon, Jinsung and Arik, Sercan and Wang, Dong and Zamani, Hamed and Han, Jiawei},
  journal={arXiv preprint arXiv:2503.09516},
  year={2025}
}

@article{wang2025ragen,
  title={Ragen: Understanding self-evolution in llm agents via multi-turn reinforcement learning},
  author={Wang, Zihan and Wang, Kangrui and Wang, Qineng and Zhang, Pingyue and Li, Linjie and Yang, Zhengyuan and Jin, Xing and Yu, Kefan and Nguyen, Minh Nhat and Liu, Licheng and others},
  journal={arXiv preprint arXiv:2504.20073},
  year={2025}
}

@article{wei2025reinforcingmultiturnreasoningllm,
  title={Reinforcing Multi-Turn Reasoning in LLM Agents via Turn-Level Reward Design}, 
  author={Quan Wei and Siliang Zeng and Chenliang Li and William Brown and Oana Frunza and Wei Deng and Anderson Schneider and Yuriy Nevmyvaka and Yang Katie Zhao and Alfredo Garcia and Mingyi Hong},
  journal={arXiv preprint arXiv:2505.11821},
  year={2025}
}

@article{dong2025agentic,
  title={Agentic reinforced policy optimization},
  author={Dong, Guanting and Mao, Hangyu and Ma, Kai and Bao, Licheng and Chen, Yifei and Wang, Zhongyuan and Chen, Zhongxia and Du, Jiazhen and Wang, Huiyang and Zhang, Fuzheng and others},
  journal={arXiv preprint arXiv:2507.19849},
  year={2025}
}

@article{aepo,
  title={Agentic entropy-balanced policy optimization},
  author={Dong, Guanting and Bao, Licheng and Wang, Zhongyuan and Zhao, Kangzhi and Li, Xiaoxi and Jin, Jiajie and Yang, Jinghan and Mao, Hangyu and Zhang, Fuzheng and Gai, Kun and others},
  journal={arXiv preprint arXiv:2510.14545},
  year={2025}
}

@article{deng2024novice,
  title={From novice to expert: Llm agent policy optimization via step-wise reinforcement learning},
  author={Deng, Zhirui and Dou, Zhicheng and Zhu, Yutao and Wen, Ji-Rong and Xiong, Ruibin and Wang, Mang and Chen, Weipeng},
  journal={arXiv preprint arXiv:2411.03817},
  year={2024}
}

@article{cui2025process,
  title={Process reinforcement through implicit rewards},
  author={Cui, Ganqu and Yuan, Lifan and Wang, Zefan and Wang, Hanbin and Zhang, Yuchen and Chen, Jiacheng and Li, Wendi and He, Bingxiang and Fan, Yuchen and Yu, Tianyu and others},
  journal={arXiv preprint arXiv:2502.01456},
  year={2025}
}

@article{liprocess,
  title={Process vs. Outcome Reward: Which is Better for Agentic RAG Reinforcement Learning},
  author={Zhang, Wenlin and Li, Xiangyang and Dong, Kuicai and Wang, Yichao and Jia, Pengyue and Li, Xiaopeng and Zhang, Yingyi and Xu, Derong and Du, Zhaocheng and Guo, Huifeng and others},
  journal={arXiv preprint arXiv:2505.14069},
  year={2025}
}

@article{zhang2025rlvmr,
  title={Rlvmr: Reinforcement learning with verifiable meta-reasoning rewards for robust long-horizon agents},
  author={Zhang, Zijing and Chen, Ziyang and Li, Mingxiao and Tu, Zhaopeng and Li, Xiaolong},
  journal={arXiv preprint arXiv:2507.22844},
  year={2025}
}

@inproceedings{yao2022react,
  title = {{ReAct}: Synergizing Reasoning and Acting in Language Models},
  author = {Yao, Shunyu and Zhao, Jeffrey and Yu, Dian and Du, Nan and Shafran, Izhak and Narasimhan, Karthik and Cao, Yuan},
  booktitle = {International Conference on Learning Representations},
  year = {2023},
  html = {https://arxiv.org/abs/2210.03629},
}

@article{achiam2023gpt,
  title={Gpt-4 technical report},
  author={Achiam, Josh and Adler, Steven and Agarwal, Sandhini and Ahmad, Lama and Akkaya, Ilge and Aleman, Florencia Leoni and Almeida, Diogo and Altenschmidt, Janko and Altman, Sam and Anadkat, Shyamal and others},
  journal={arXiv preprint arXiv:2303.08774},
  year={2023}
}

@article{wei2022chain,
  title={Chain-of-thought prompting elicits reasoning in large language models},
  author={Wei, Jason and Wang, Xuezhi and Schuurmans, Dale and Bosma, Maarten and Xia, Fei and Chi, Ed and Le, Quoc V and Zhou, Denny and others},
  journal={Advances in Neural Information Processing Systems},
  volume={35},
  pages={24824--24837},
  year={2022}
}

@article{singhal2023large,
  title={Large language models encode clinical knowledge},
  author={Singhal, Karan and Azizi, Shekoofeh and Tu, Tao and Mahdavi, S Sara and Wei, Jason and Chung, Hyung Won and Scales, Nathan and Tanwani, Ajay and Cole-Lewis, Heather and Pfohl, Stephen and others},
  journal={Nature},
  volume={620},
  number={7972},
  pages={172--180},
  year={2023},
  publisher={Nature Publishing Group}
}

@article{wu2023bloomberggpt,
  title={Bloomberggpt: A large language model for finance},
  author={Wu, Shijie and Irsoy, Ozan and Lu, Steven and Dabravolski, Vadim and Dredze, Mark and Gehrmann, Sebastian and Kambadur, Prabhanjan and Rosenberg, David and Mann, Gideon},
  journal={arXiv preprint arXiv:2303.17564},
  year={2023}
}

@inproceedings{
qin2025learnropestrustwins,
title={Learn the Ropes, Then Trust the Wins: Self-imitation with Progressive Exploration for Agentic Reinforcement Learning},
author={Yulei Qin and Xiaoyu Tan and Zhengbao He and Gang Li and Haojia Lin and Zongyi Li and Zihan Xu and Yuchen Shi and Siqi Cai and Renting Rui and Shaofei Cai and Yuzheng Cai and Xuan Zhang and Sheng Ye and Ke Li and Xing Sun},
booktitle={International Conference on Learning Representations},
year={2026},
url={https://openreview.net/forum?id=Kssko33Ekq}
}

@article{bai2022constitutional,
  title={Constitutional ai: Harmlessness from ai feedback},
  author={Bai, Yuntao and Kadavath, Saurav and Kundu, Sandipan and Askell, Amanda and Kernion, Jackson and Jones, Andy and Chen, Anna and Goldie, Anna and Mirhoseini, Azalia and McKinnon, Cameron and others},
  journal={arXiv preprint arXiv:2212.08073},
  year={2022}
}

@inproceedings{oh2018self,
  title={Self-imitation learning},
  author={Oh, Junhyuk and Guo, Yijie and Singh, Satinder and Lee, Honglak},
  booktitle={International Conference on Machine Learning},
  pages={3878--3887},
  year={2018},
  organization={PMLR}
}

@article{amodei2016concrete,
  title={Concrete problems in AI safety},
  author={Amodei, Dario and Olah, Chris and Steinhardt, Jacob and Christiano, Paul and Schulman, John and Man{\'e}, Dan},
  journal={arXiv preprint arXiv:1606.06565},
  year={2016}
}

@article{schulman2016high,
  title={High-dimensional continuous control using generalized advantage estimation},
  author={Schulman, John and Moritz, Philipp and Levine, Sergey and Jordan, Michael and Abbeel, Pieter},
  journal={arXiv preprint arXiv:1506.02438},
  year={2015}
}

@article{zhang2025agent,
  title={Agent learning via early experience},
  author={Zhang, Kai and Chen, Xiangchao and Liu, Bo and Xue, Tianci and Liao, Zeyi and Liu, Zhihan and Wang, Xiyao and Ning, Yuting and Chen, Zhaorun and Fu, Xiaohan and others},
  journal={arXiv preprint arXiv:2510.08558},
  year={2025}
}

@article{gao2025soft,
  title={Soft adaptive policy optimization},
  author={Gao, Chang and Zheng, Chujie and Chen, Xiong-Hui and Dang, Kai and Liu, Shixuan and Yu, Bowen and Yang, An and Bai, Shuai and Zhou, Jingren and Lin, Junyang},
  journal={arXiv preprint arXiv:2511.20347},
  year={2025}
}

@article{igpo,
  title={Information Gain-based Policy Optimization: A Simple and Effective Approach for Multi-Turn LLM Agents},
  author={Wang, Guoqing and Dai, Sunhao and Ye, Guangze and Gan, Zeyu and Yao, Wei and Deng, Yong and Wu, Xiaofeng and Ying, Zhenzhe},
  journal={arXiv preprint arXiv:2510.14967},
  year={2025}
}


\newpage

\newpage

\appendix

\section{A~~~Implementation Details}\label{details}

All our experiments are implemented based on VeRL-Agent.~\cite{gigpo}. Throughout our experiments, we strictly adhered to the licensing terms for academic use associated with LLaMA-3.2-3B-Instruct. The licensing terms can be found at the following link: \url{https://huggingface.co/meta-llama/Llama-3.2-3B/blob/main/LICENSE.txt}.



\subsection{Hyperparameter Settings}

To ensure a fair comparison, we follow the settings of GiGPO and retained most of the primary parameters from VeRL-Agent. For the additional hyperparameters introduced, the off-policy data ratio $\alpha$ is set to $1/8$. For GRPO and GiGPO, the train data size is set to $16$. For RSPO-based methods, the train data size comprises $14$ on-policy and $2$ off-policy data size. Consequently, the total batch size for PPO is $128 (16 \times 8)$, and for RSPO-PPO is $112 (14 \times 8)$.

In each RSPO loop, we trained the agent for $k$ steps using dense rewards, followed by $k$ steps using outcome rewards, where $k=3$ (the total step count is calculated based on the update steps of the final obtained model). When populating the replay buffer, the training data size was set to $14 \times 2 = 28$, and the rollout group size $N=8$. All experiments run for a total of $150$ steps (counted based on the update steps of the final obtained model).

\subsection{Computing resources}\label{compute}

All experiments are conducted on workstations equipped with 380 CPU cores, 2.2TB memory, and 8 GPUs with 96GB memory. In WebShop, the 1.5B and 7B models are trained on 2 and 8 GPUs respectively for GRPO and GiGPO, whereas PPO required 4 and 8 GPUs respectively. In ALFWorld, the 1.5B model is trained on 8 GPUs, while the 7B model uses 2 workstations with 16 GPUs.

\section{B~~~Hyperparameter Sensitivity Analysis}\label{hyper}
We conduct additional ablation studies on the WebShop using Qwen2.5-1.5B-Instruct model. To ensure a strictly fair comparison across all hyperparameter sweeps, we re-run the baseline and all variants under a unified experimental setting (e.g., consistent random seeds and compute budget). While absolute scores vary slightly due to seed variance, the relative performance gains and trends remain consistent with the main paper.

\subsection{Sensitivity Analysis of Off-policy Data Ratio $\alpha$}

We evaluated $\alpha \in \{1/16, 1/8, 1/4, 1/2\}$ while fixing $k=3$. The results are compared against the GRPO baseline under the same settings.

\begin{table}
  \centering
  \setlength{\tabcolsep}{1.4mm}
  \begin{tabular}{llcc}
        \toprule
        Method & Setting & Score & Success Rate (SR) \\
        \midrule
        GRPO & - & 82.5 & 61.7 \\
        \midrule
        RSPO & $\alpha=1/16$ & \textbf{85.1} & \textbf{67.2} \\
        RSPO & $\alpha=1/8$ (Default) & 85.0 & 66.4 \\
        RSPO & $\alpha=1/4$ & 77.4 & 54.7 \\
        RSPO & $\alpha=1/2$ & 79.1 & 60.9 \\
        \bottomrule
    \end{tabular}
    \caption{Ablation study of $\alpha$ in RSPO (with fixed $k=3$)}
  \label{tab:alpha}
\end{table}


When $\alpha$ is kept small (e.g., $1/16$ to $1/8$), the method effectively leverages diverse trajectories to improve performance without disrupting training stability. As $\alpha$ increases to $1/4$ or $1/2$, performance drops significantly. This aligns with our discussion in Section 4.2: excessive off-policy data introduces severe distribution shifts, which destabilizes the policy update.

\subsection{Sensitivity Analysis of Loop Steps $k$}

We evaluated $k \in \{1, 3, 5, 10\}$ while fixing $\alpha=1/8$.

\begin{table}[htp]
  \centering
  \setlength{\tabcolsep}{1.4mm}
    \begin{tabular}{llcc}
        \toprule
        Method & Setting & Score & Success Rate (SR) \\
        \midrule
        GRPO & - & 82.5 & 61.7 \\
        \midrule
        RSPO & $k=1$ & 81.0 & 65.6 \\
        RSPO & $k=3$ (Default) & \textbf{85.0} & \textbf{66.4} \\
        RSPO & $k=5$ & 78.5 & 61.7 \\
        RSPO & $k=10$ & 82.3 & 61.7 \\
        \bottomrule
    \end{tabular}
      \caption{Ablation study of $k$ in RSPO (with fixed $\alpha=1/8$).}
  \label{tab:k}
\end{table}


We observe an "inverted-U" trend, indicating that $k$ must be balanced to avoid under-exploration or overfitting to dense rewards. When $k=3$, this setting achieves the best balance, yielding the highest Score and Success Rate (SR). When $k=1$, the performance is slightly lower compared to the optimal setting. And when $k \ge 5$, the performance reverts to the baseline level. Optimizing against the dense reward model for too long may lead to signs of reward hacking, resulting in no improvement in final performance.

These experiments confirm that RSPO is robust and effective as long as hyperparameters are selected within a reasonable range (i.e., a small $\alpha$ to maintain stability and a moderate $k$ to balance exploration).

\definecolor{printblue}{cmyk}{1.0, 0.5, 0.0, 0.0}
\begin{algorithm*}[t]
    \caption{Training Agentic LLMs with RSPO}
    \label{alg:code}
        \begin{algorithmic}[1]
        \STATE {\bfseries Require:} Initial model $\pi_{\text{base}}$, on-policy proximal PG method $\mathbb{P}$ (e.g., PPO, GRPO), dense reward model $R$, task distribution $p(X)$, group size $N$, replay buffer $\mathcal{D}$, training batch size $B$, training steps $k$ per loop,
        off-policy data proportion $\alpha$.
        \STATE Init the model $\pi_{\theta} = \pi_{\text{base}}$ and $\mathcal{D}=\emptyset$
        \FOR{each loop of RSPO}
            \STATE {\color{printblue}{\# Update using dense rewards}}
            \STATE Set the model $\pi_{dense} = \pi_{\theta}$
            \FOR{$\text{step} = 1, \ldots, k$}
                \STATE Sample task $x \sim p(X)$ with size $(1-\alpha)B$ and generate trajectories according to $\pi_{dense}$ and $\mathbb{P}$
                \STATE Compute dense rewards for trajectories using R
                \STATE Update $\pi_{dense}$ using dense rewards according to $\mathbb{P}$
            \ENDFOR
            \STATE Collect trajectories $\tau$ using $\pi_{dense}$, re-evaluate with outcome rewards, and add to replay buffer $\mathcal{D}$
            \STATE {\color{printblue}{\# Update using outcome rewards}}
            \FOR{$\text{step} = 1, \ldots, k$}
                \STATE Sample task $x \sim p(X)$ with size $(1-\alpha)B$ and generate trajectories $\mathcal{D}_{on}$ according to $\pi_{\theta}$
                \STATE Sample trajectories $\mathcal{D}_{off}$ with size $\alpha BN$ from replay buffer $\mathcal{D}$
                \STATE $\mathcal{D}_{mix} \leftarrow \mathcal{D}_{on} \cup \mathcal{D}_{off}$
                \STATE Update $\pi_{\theta}$ using outcome rewards and $\mathcal{D}_{mix}$ according to $\mathbb{P}$ (Eq.~1)
            \ENDFOR
            \STATE Reset the replay buffer $\mathcal{D}\leftarrow\emptyset$
        \ENDFOR
        \RETURN $\pi_{\theta}$
        \end{algorithmic}
\end{algorithm*}

\section{C~~~Sensitivity to Dense Reward Quality}\label{noise}

To investigate the sensitivity of RSPO to the quality of dense rewards, we conducted an ablation study on WebShop using Qwen2.5-1.5B-Instruct. Specifically, we introduced Gaussian noise $\mathcal{N}(0, \sigma)$ to the dense rewards used by the agent during exploration.

\begin{table}[htp]
  \centering
  \setlength{\tabcolsep}{0.5mm}
    \begin{tabular}{lccc}
        \toprule
        Method & Noise Level ($\sigma$) & Score & Success Rate (SR) \\
        \midrule
        RSPO (Standard) & 0 & \textbf{85.0} & \textbf{66.4} \\
        RSPO + Noise & 0.1 & 82.1 & 64.1 \\
        RSPO + Noise & 0.3 & 81.1 & 60.9 \\
        \bottomrule
    \end{tabular}
      \caption{Impact of dense reward noise $\sigma$ on RSPO performance.}
  \label{tab:noise}
\end{table}

As shown in the table, when the noise is relatively small ($\sigma=0.1$), the performance remains robust (SR: 66.4 vs 64.1), demonstrating that RSPO does not require high-precision dense rewards to be effective.
Even with significant noise ($\sigma=0.3$), the performance drop is moderate, and the model does not collapse.
This validates our core insight: the dense reward in RSPO serves primarily as a "guide" for exploration rather than a ground truth. As noted in Figure 3 of our paper, training solely with the dense reward (even without added noise) leads to reward hacking. RSPO succeeds because the "Reward-Swap" mechanism uses the true outcome reward to filter out trajectories, ensuring that even if the dense reward is imperfect or noisy, the final policy remains aligned with the task objective.





\section{D~~~Pseudo Code}
\label{sec:appendix_code}

Algorithm~\ref{alg:code} summarizes the full training procedure of RSPO. Each RSPO loop consists of two distinct training phases. In the first phase, the model $\pi_{dense}$ is trained for $k$ steps using dense rewards. The trained $\pi_{dense}$ then interacts with the environment to generate a batch of trajectories, which are labeled with outcome rewards and stored in the replay buffer $\mathcal{D}$. 
Subsequently, the process enters the second phase, where the model $\pi_{\theta}$ is trained for $k$ steps using outcome rewards. The training dataset employed here is a mixture of on-policy data generated by $\pi_{\theta}$ and off-policy data sampled from the replay buffer.
This cycle is repeated to continuously refine $\pi_{\theta}$ and $\pi_{dense}$.

This training paradigm ensures that the final model $\pi_{\theta}$ is consistently trained with task outcome rewards, thereby avoiding "Reward Hacking" issues stemming from potential inconsistencies in dense process rewards. Simultaneously, it expands the model's exploration space by leveraging data from the replay buffer, ultimately elevating the model's performance ceiling.

\section{E~~~Limitations}\label{limit}

Due to computational resource constraints, we did not conduct experiments on models exceeding 7B parameters; consequently, the performance gains of RSPO on larger-scale models remain to be fully validated. Additionally, the off-policy data ratio is kept constant throughout our experiments. Employing an adaptive ratio (e.g., a decaying schedule) could potentially enhance the final results. We identify these as promising directions for future work.

\end{document}